%% file: main.tex
\documentclass[preprint]{IEEEtran}
\IEEEoverridecommandlockouts

\usepackage{caption}
\usepackage{cite}
\usepackage{xspace}
\usepackage{amsmath,amssymb,amsfonts}
\usepackage{algorithmic}
\usepackage{graphicx}
\usepackage{multirow}
\usepackage{textcomp}
\usepackage{xcolor}
\usepackage{balance}
\usepackage{hyperref}
\usepackage{subcaption}
\newcommand{\ie}{\textit{i.e.}\xspace}
\newcommand{\eg}{\textit{e.g.}\xspace}

\newcommand{\etal}{\textit{et al.}\xspace}

\def\BibTeX{{\rm B\kern-.05em{\sc i\kern-.025em b}\kern-.08em
    T\kern-.1667em\lower.7ex\hbox{E}\kern-.125emX}}

\begin{document}

\title{\emph{Images in Motion?}: A First Look into Video Leakage in Collaborative Deep Learning}
\author{
\IEEEauthorblockN{Md Fazle Rasul\IEEEauthorrefmark{1}, Alanood Alqobaisi\IEEEauthorrefmark{1}, Bruhadeshwar Bezawada\IEEEauthorrefmark{2} and Indrakshi Ray\IEEEauthorrefmark{1}\thanks{This work was partially supported by NSF under Grant No. CNS 1822118, CNS 2226232,  CNS 2335687, DMS 2123761, by the industry member partners of the NSF IUCRC Center for Cyber Security Analytics and
Automation (AMI, NewPush, Cyber Risk Research, NIST and ARL), by the State of Colorado (grant \#SB 18-086), by NIST Grant No. 60NANB23D152 and by the
authors’ institutions.}
} \\
\IEEEauthorblockA{\IEEEauthorrefmark{1}Department of Computer Science\\
Colorado State University, Fort Collins, CO 80523-1873\\
Email: \{Fazle.Rasul,Alanood.Alqobaisi,Indrakshi.Ray\}@colostate.edu}
\IEEEauthorblockA{\IEEEauthorrefmark{2}Department of Mathematics and Computer Science, Southern Arkansas University, Magnolia, AR 71753-5000
\\
Email: bbezawada@saumag.edu}
}

\maketitle

\begin{abstract}
Federated learning (FL) allows multiple entities to train a shared model collaboratively. 
Its core, privacy-preserving principle is that participants only exchange model updates, such as gradients, and never their raw, sensitive data. 
This approach is fundamental for applications in domains where privacy and confidentiality are important. 
However, the security of this very mechanism is threatened by gradient inversion attacks, which can reverse-engineer private training data directly from the shared gradients, defeating the purpose of FL. 
While the impact of these attacks is known for image, text, and tabular data, their effect on video data remains an unexamined area of research. 
This paper presents the first analysis of video data leakage in FL using gradient inversion attacks. 
We evaluate two common video classification approaches: one employing pre-trained feature extractors and another that processes raw video frames with simple transformations. 
Our initial results indicate that the use of feature extractors offers greater resilience against gradient inversion attacks.

We also demonstrate that image super-resolution techniques can enhance the frames extracted through gradient inversion attacks, enabling attackers to reconstruct higher-quality videos. 
Our experiments validate this across scenarios where the attacker has access to zero, one, or more reference frames from the target environment. 
We find that although feature extractors make attacks more challenging, leakage is still possible if the classifier lacks sufficient complexity.
We, therefore, conclude that video data leakage in FL is a viable threat, and the conditions under which it occurs warrant further investigation.
\end{abstract}

\sloppy
\input{intro}
\input{problem_statement}
\input{related}

\input{proposed}
\input{performance}
\input{discussion}
\input{conclusion}
\bibliographystyle{plain}
\bibliography{references}
\end{document}

%% file: intro.tex
\section{Introduction}
\label{sec:intro}
Many domains, including healthcare, law enforcement, and national forest services, rely on analyzing sensitive video data to detect anomalous events.
Machine learning appears to be a promising approach for anomaly detection. 
However, a single organization may not have enough volume of data, especially pertaining to various types of anomalies,  to build robust machine learning models for detecting 
 different anomalous behavior. 
 This limitation necessitates training collaboratively across different organizations.
 Sharing sensitive data is often undesirable 
as it may reveal confidential information about the organization and the subjects contained in them; data sharing without consent may even be illegal because of data privacy regulations, including General Data Protection Regulation (GDPR), the California Consumer Privacy Act (CCPA), and the Australian Privacy Principles (APP). 

Federated Learning (FL) \cite{kingma2014adam,2019amsgrad, reddi2019convergence} appears 
to be a feasible solution as it allows multiple parties to construct a global model by sharing the gradients of the local models, where the gradients represent the training errors on the local data and are used to adjust the weights in the global machine learning model.
As these algorithms do not require the sharing of local sensitive data, they are considered privacy-preserving.
However, several recent attacks \cite{zhu2019deep, zhao2020idlg, geiping2020inverting, zhu2020r, wang2020sapag, lam2021gradient, wei2020framework, yin2021see, jeon2021gradient, jin2021cafe} 
have shown that federated learning is susceptible to training data leakage from the shared training gradients. 
We investigate the problem of deep leakage attacks in video data, a process enabling the full reconstruction of raw samples from model gradients.

Much of the existing literature has focused on deep leakage attacks against image, text, and tabular data, but not on video data.
Leakage attacks against image classification systems have appeared in \cite{zhu2019deep, yin2021see, 
zhao2020idlg, geiping2020inverting, li2021deep, ding2024improved, du2024sok,zhu2020r}.
Similar extensions of attacks have been described for tabular \cite{vero2022tableak} and vertical data \cite{jin2021cafe}.
Zhu \etal \cite{zhu2019deep} investigated
``shallow" leakage of video data; shallow leakage reveals only general data properties like membership information or aggregate properties of the dataset, in contrast to deep leakage, which exposes the data itself.
Our focus is to detect deep leakage attacks on video classification systems, such as the video captioning system from \cite{wu2017deep}  and the more recent collaborative anomaly detection system in \cite{al2024collaborative} that detects intruders using security camera feeds.

Video data presents its unique characteristics in the context of machine learning, primarily due to its temporal dimension. 
This time-dependent structure of videos introduces two fundamental issues that require careful consideration.
First, it drastically increases data volume compared to images. 
Second, it creates high inter-frame temporal redundancy, as consecutive frames contain substantial overlapping information. 
This is a security vulnerability, as adversaries can exploit it to reconstruct a high-fidelity version of the original video from only a small number of frames.
%

%
Training on video data broadly adopts two different approaches to handle the temporal sequence of frames in videos. 
The first and straightforward approach treats a video as a collection of frames, making the video data set functionally equivalent to a static image dataset. 
The second and popular approach acknowledges the temporal sequence of videos. 
In this case, a video is segmented into clips, and a pre-trained convolutional neural network, \eg ResNet \cite{resnet}, AlexNet \cite{alexnet}, is used to extract deep features from the frames. 
These features are complex representations of the frame content learned by the neural network's multiple layers.

These frame-level features are then aggregated, often through averaging, to form a single, comprehensive video-level representation for each clip. 
These features are then used for classification \cite{wu2017deep, al2024collaborative}. 
This aggregation step is a key differentiator between video processing and image processing. 
When feature extractors are used for the image data set, each image typically retains its own distinct feature matrix.

Current studies \cite{zhu2019deep, zhao2020idlg, geiping2020inverting, zhu2020r, yin2021see}
on gradient inversion attacks that demonstrate successful image reconstruction, typically assume that gradients are derived directly from input images and not from the features of the input images.  
Our work covers a more challenging scenario where classifiers are trained on extracted features. 
We investigated whether such training, which is based on intermediate data representations, mitigates the risks associated with attacks based on gradient inversion. 
Thus, we are the first to demonstrate the impact of gradient inversion attacks on extracted features.

We first explored whether current gradient inversion attacks can be effectively extended to reconstruct meaningful information from complex video data under the previously mentioned scenarios: gradients computed on features generated by a pre-trained feature extractor and gradients computed on raw video frames with simple transformations.
We used the deep leakage from gradients (DLG) attack described in \cite{zhu2019deep} to reconstruct the frames.
However, since the original attack in \cite{zhu2019deep} is intended to reconstruct images under a certain resolution, we had to make adjustments to the video frames' resolution.
But the low-resolution frames extracted using the DLG attack result in visually degraded video outputs.
To overcome this, we propose the use of image super-resolution algorithms \cite{wang2021realesrgan, yang2020learning} to improve the quality of the extracted video frames. 
We simulated scenarios where the attacker has access to zero, one, or more sample reference video frames from the target environment, which assists in the reconstruction of the neighboring video frames.
Finally, we investigated the impact of feature extractors on gradient inversion attacks, conducting the analysis on an image dataset where these attacks are demonstrably successful.
The main contributions of our work are as follows:
\begin{itemize}
\item To the best of our knowledge,  we are the first to explore deep leakage from gradient inversion attacks on video data. 
%
%
\item Furthermore, we demonstrate how the quality of frames generated by the gradient inversion attacks can be improved using video resolution enhancement techniques, both with and without sample reference frames. 
We conclude that, video enhancement techniques used in conjunction with the gradient inversion attacks help to exacerbate the impact of the attack.
%
%
\item We demonstrate, on the UCF-Crime \cite{sultani2018real} dataset, that employing a pre-trained feature extractor makes the reconstruction of individual video frames from gradients much more challenging, establishing its effectiveness as a practical defense.
\item We present a systematic analysis, using the CIFAR-100 dataset \cite{krizhevsky2009learning}, that isolates and quantifies the defensive impact of feature extractors against gradient inversion attacks, providing generalizable insights into the factors that enhance privacy.
\end{itemize}

\noindent{\bf Organization. }. Section \ref{sec:problem_statement} presents the problem statement. We establish the context of our solution in Section \ref{sec:prior} by reviewing the landscape of existing gradient inversion attacks and defenses. Section \ref{sec:proposed} provides details of our proposed approach. Section \ref{sec:performance} rigorously evaluates our methods, presenting comprehensive experimental results.  Section \ref{sec:discussion} explores the open challenges. Finally, Section \ref{sec:conclusion} concludes the paper and discusses future work.

%% file: problem_statement.tex
\section{Problem Statement}
\label{sec:problem_statement}
%
We consider that $n$ parties,
denoted by $\{P_1, P_2, \ldots, P_n\}$,
are participating in the collaborative learning algorithm  and possess  $n$ distinct sets of data, represented by, $\{D_1, D_2, \ldots, D_n\}$, respectively.
Each $D_i$, $\forall i  \leq n$, consists of $M$ video data frames of certain dimensions and each video frame is denoted by $V^i_j$ where $j \leq M$.
We denote the local machine learning model for $P_i$ as $C_i$ and the number of training epochs per data sample by $t$.
Let $\delta^k (V^i_j)$, where $k \leq t$, denote the gradients generated by training $C_i$ on $V^i_j$ in the $k^{th}$ epoch.
At the end of each training epoch, these gradients are shared by each participant with the other participants. 

The problem is thus: Given $\delta^k (V^i_j)$ where $i \neq x$ and $\forall k \leq t$, a participant $P_x$ needs an algorithm to reconstruct $V^i_j$ without any additional information from $P_i$.
%
%
Additionally, we assume that $P_x$ is a semi-honest participant, \ie, $P_x$ follows the rules of the collaborative protocol but will attempt to learn the private training data of the other participants.

%% file: related.tex
\section{Related Work}
\label{sec:prior}
We first discuss the methodologies and evolution of gradient inversion attacks, and then provide a comprehensive analysis of the strategies proposed to mitigate them.
\subsection{Gradient Leakage Attacks on Federated Learning}
The vulnerability of gradients was starkly exposed in seminal works that formulated data reconstruction as an optimization problem. 
The DLG paper was a pioneering effort \cite{zhu2019deep}, demonstrating that by initializing a dummy input and label, an attacker could iteratively optimize them to produce a gradient that closely matches the target gradient, thus recovering the original data and label pixel-for-pixel. 
%
%
A subsequent work, iDLG \cite{zhao2020idlg}, enhanced this process by showing that ground-truth labels could be extracted analytically from the gradients of a model's final fully-connected layer, which significantly accelerated the image reconstruction process by removing the need to optimize the label.

Concurrently, researchers in \cite{geiping2020inverting} framed the attack as minimizing the cosine distance between gradients rather than the Euclidean distance, which proved more robust to variations in gradient magnitude. 
This work also introduced total variation as a regularization term to improve the visual quality of reconstructed images, demonstrating successful recovery even from batches of images. 
Further solidifying these findings, the GradInversion attack \cite{yin2021see} demonstrated robust image batch recovery and introduced an effective method for recovering labels from a batch even when multiple classes are present. 
These early works collectively established that gradients, far from being secure, contain substantial, extractable information about the training data. 
The core principle of these attacks was further analyzed in \cite{chen2021understanding}, which provided a layer-wise theoretical understanding of how architectural properties contribute to data leakage.

%
One line of work introduced powerful priors to guide the reconstruction process. 
Jeon \etal  \cite{jeon2021gradient} showed that employing a pre-trained generative model could produce significantly more realistic and faithful reconstructions, as the optimization is performed in the generator's latent space rather than the high-dimensional pixel space. 
This marked a major leap in the quality of recovered data, moving from noisy or blurry images to highly plausible ones.

%
Further, Zhu \etal  \cite{zhu2020r} proposed a novel analytical approach that recursively solves for layer inputs in a closed-form manner, offering a much faster alternative to iterative optimization and providing insights into which network architectures are inherently more vulnerable due to rank deficiency. 
The issue of applicability was tackled by demonstrating that these attacks are not limited to standard computer vision tasks but also pose a severe threat to sensitive domains, as shown in \cite{li2021deep}, which successfully reconstructed multi-label medical images.
The survey by Wang \etal. \cite{wang2023beyond} provides a more advanced attack vector involving transferability, where an attack on one target proves effective against another.

As the FL field matured, so did the attacks designed to break it. 
Recognizing that defenses like dropout were being considered, the work \cite{scheliga2023dropout} demonstrated that while standard inversion attacks fail against dropout's stochasticity, a modified attack that jointly optimizes for both the private data and the dropout mask can successfully bypass this defense. 
Similarly, to address the common use of gradient compression to save bandwidth, researchers developed an improved attack in \cite{ding2024improved} showing that even heavily compressed or pruned gradients could be vulnerable.

The scope of attacks also broadened beyond the standard horizontal FL setting. 
Jin \etal \cite{jin2021cafe} revealed that vertical federated learning (VFL), where clients hold different features of the same data samples, is particularly vulnerable. 
In VFL, since the server often knows the indices of data samples used in a batch, this side-channel information can be exploited to drastically improve reconstruction, even for very large batches.
Furthermore, the problem of attacking aggregated gradients from multiple users was explored in \cite{lam2021gradient}, which showed that by observing aggregated updates over several rounds and leveraging side-channel information about user participation, an attacker could potentially disentangle individual user updates, thereby enabling traditional gradient inversion on the disaggregated gradients.

Finally, frameworks have been proposed to formalize and evaluate these privacy risks. 
Balunović \etal \cite{balunovic2021bayesian} established a theoretical basis for understanding the problem by defining a Bayes optimal adversary, allowing existing attacks to be interpreted as approximations of this theoretical ideal. 
Other works, such as \cite{du2024sok}, and evaluation frameworks like those in \cite{huang2021evaluating} provide a comprehensive overview and structured analysis of the attack landscape, categorizing existing methods and critically assessing their effectiveness under practical conditions. 
Together, this body of literature confirms that gradient inversion is a persistent and evolving threat that challenges the core privacy promises of federated learning.
\subsection{Defenses Against Training Data Leakage} 
The defense strategies can be broadly categorized into three main categories: perturbing or modifying gradients, obfuscating data before training, and employing cryptographic techniques to protect gradient information.

A foundational defense strategy involves perturbing the gradients before they are shared. 
The most formally recognized approach in this category is Differential Privacy, which adds precisely calibrated noise to the gradients. 
As detailed in \cite{abadi2016deep}, this method provides a provable guarantee that the inclusion of any single data point in the training set has a limited and quantifiable impact on the output, thereby making accurate data reconstruction from the noisy gradients computationally infeasible. 
Other perturbation techniques focus on information reduction. Revisiting Gradient Pruning explores the sparsification of gradients by setting a significant fraction of the values with the smallest magnitudes to zero. 
This intentional removal of information degrades the quality of the gradient signal available to an attacker \cite{xue2024revisiting}. 
A more advanced gradient manipulation technique is proposed in \cite{zhang2025censor}, which defends against attacks by projecting the true gradient onto a randomly sampled orthogonal subspace. 
This censoring process is designed to remove sufficient information to thwart reconstruction while preserving enough to maintain model utility.

Another class of defenses operates on the data itself, before any gradients are computed. 
The core idea is to make the relationship between the shared gradient and any single, original data sample ambiguous. 
The mixup framework, introduced in \cite{zhang2017mixup}, achieves this by training the model on convex combinations of pairs of data samples and their labels. 
The resulting gradients correspond to these virtual, mixed samples, not the original ones, confounding reconstruction attempts. 
Taking this concept further, \cite{huang2020instahide} proposes a more complex data-hiding scheme where each private training image is mixed with several images from a public dataset using a unique invertible mixing matrix. 
This defense places a significant computational burden on the attacker, who would need to solve a difficult disentanglement problem to recover the original private sample.

Finally, cryptographic methods offer powerful protection by preventing the server from ever observing individual client gradients. Secure Aggregation protocols, outlined in works like \cite{bonawitz2016practical}, ensure that the central server only receives the sum of gradients from a cohort of users, not the individual contributions. 
This makes it impossible to isolate and attack a single user's update.
Recognizing the potential communication and computation overhead of such methods, subsequent research, such as FastSecAgg \cite{kadhe2020fastsecagg} and LightSecAgg \cite{yang2021lightsecagg}, has focused on developing more efficient and scalable secure aggregation protocols. 
An alternative cryptographic approach, discussed in Privacy-Preserving Deep Learning via additively homomorphic encryption \cite{aono2017privacy}, uses encryption schemes that allow the server to compute the sum of encrypted gradients without decrypting them, achieving a similar privacy guarantee. 
%
%
For instance, in \cite{guo2025new}, the authors propose a two-phase training process that aims to break the invertible relationship between gradients and private data by creating a more complex and indirect mapping. 
%
%
%
While these works demonstrate gradient inversion attacks and defenses on static data like images, these methods do not account for the unique challenges of video. 
Our research bridges this critical gap by providing the first systematic investigation into gradient inversion attacks on video data, analyzing both the feasibility of the attack and the efficacy of feature extractors as a practical defense.

%% file: proposed.tex
\section{Proposed Approach}
\label{sec:proposed}
This section details our multi-stage pipeline for video reconstruction and enhancement, beginning with an architectural overview before elaborating on each stage.
\subsection{Architectural Overview}
\label{subsec:overview}
\begin{figure}[!t]
    \includegraphics[width = 9 cm]{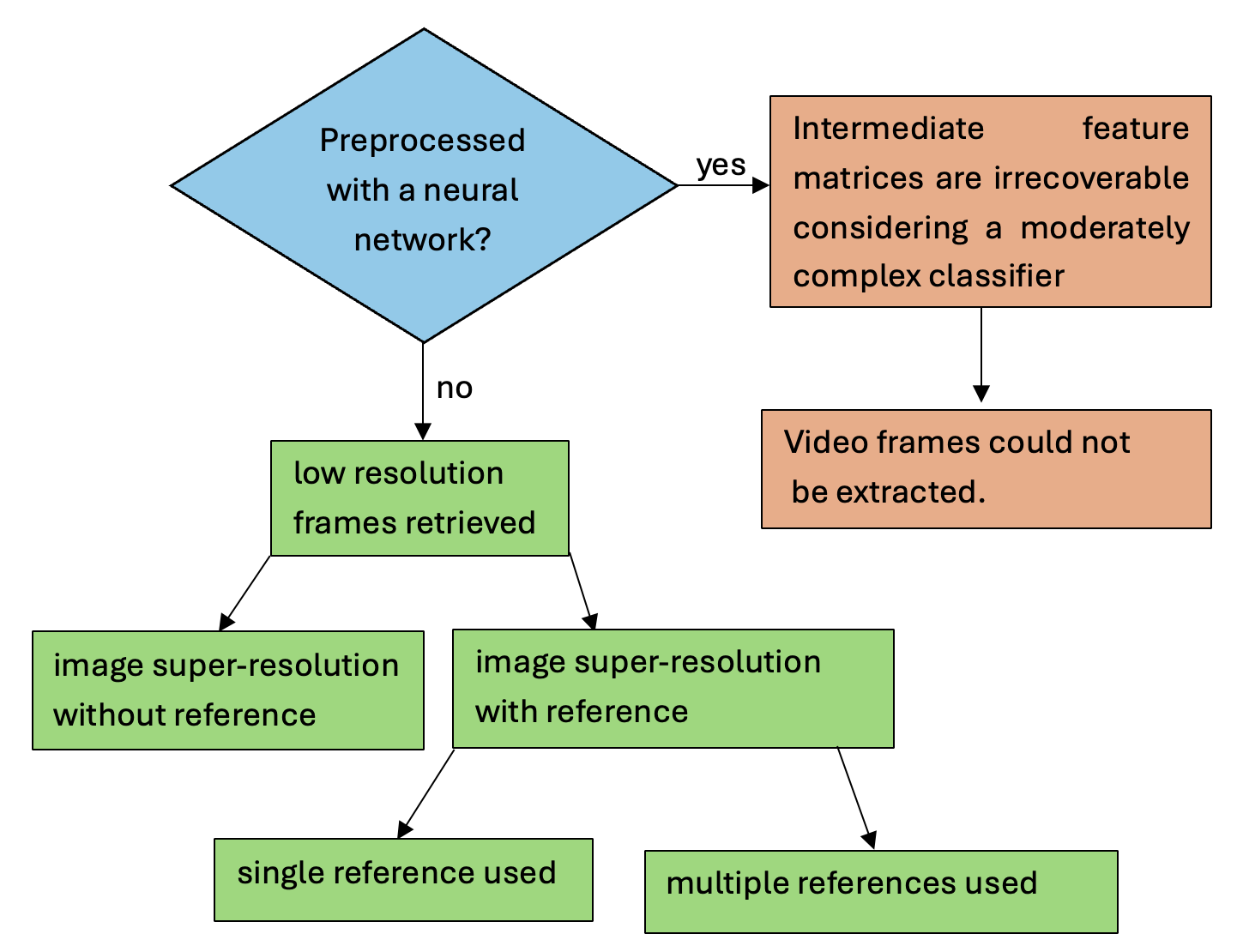}
    \caption{Video Processing Pipeline}
    \label{fig:pipeline}
\end{figure}

We have rigorously examined potential data leakage pathways in video-based training, exploring various handling methodologies.
Furthermore, a detailed analysis of the quality and utility of any leaked data was performed, specifically assessing its value to a potential adversary and ways in which the quality of the extracted frames can be improved. 
Fig. \ref{fig:pipeline} illustrates the video processing pipeline explored in this study. 
We investigated the impact of neural network pre-processing on video frame retrieval. 
As depicted in Fig. \ref{fig:pipeline}, the pipeline branches according to whether a neural network is used for initial processing. 
When pre-processing was used, we observed that intermediate feature representations, crucial for frame reconstruction, were not retrievable when a moderately complex classifier was used, preventing the reconstruction of the original video frames.
Section \ref{subsec:features} describes this case.  
Section \ref{subsec:raw_video_frames} explores cases where 
pre-processing is not applied, and low-resolution frames are extracted.
These frames are then subjected to super-resolution. 
Our simulations evaluated the attacker's capabilities across three scenarios: no reference frame access, single reference frame access, and multiple reference frame access, providing insights into the impact of varying levels of attacker capabilities. 
Finally, in section \ref{subsec:effect_of_using_features} we will discuss our methods for evaluating the effects of feature extractors on gradient inversion attacks.
%
%
\subsection{Gradient Inversion on Raw Frames with Simple Transformation}
\label{subsec:raw_video_frames}
We first explored video reconstruction in the absence of neural network feature extractors, simulating cases where videos undergo only nominal transformations like rotation, resizing, cropping, normalizing, etc. 
Using the DLG attack \cite{zhu2019deep}, detailed in section \ref{subsec:effect_of_using_features}, we extracted the video frames from the shared gradients. 
Then we implemented methodologies from existing literature to improve the fidelity of reconstructed frames. 
Existing research \cite{li2021deep} demonstrates the feasibility of gradient inversion attacks on deep neural networks, successfully reconstructing images from gradients with batch sizes ranging from 8 to 48 and image dimensions up to $224$x$224$ pixels. 
So, instead of emphasizing batch sizes and image dimensions, our research specifically investigated methods for improving the fidelity of extracted video frames. 
\begin{figure*}[htbp]
    \centering
    \includegraphics[width = 0.7\textwidth]{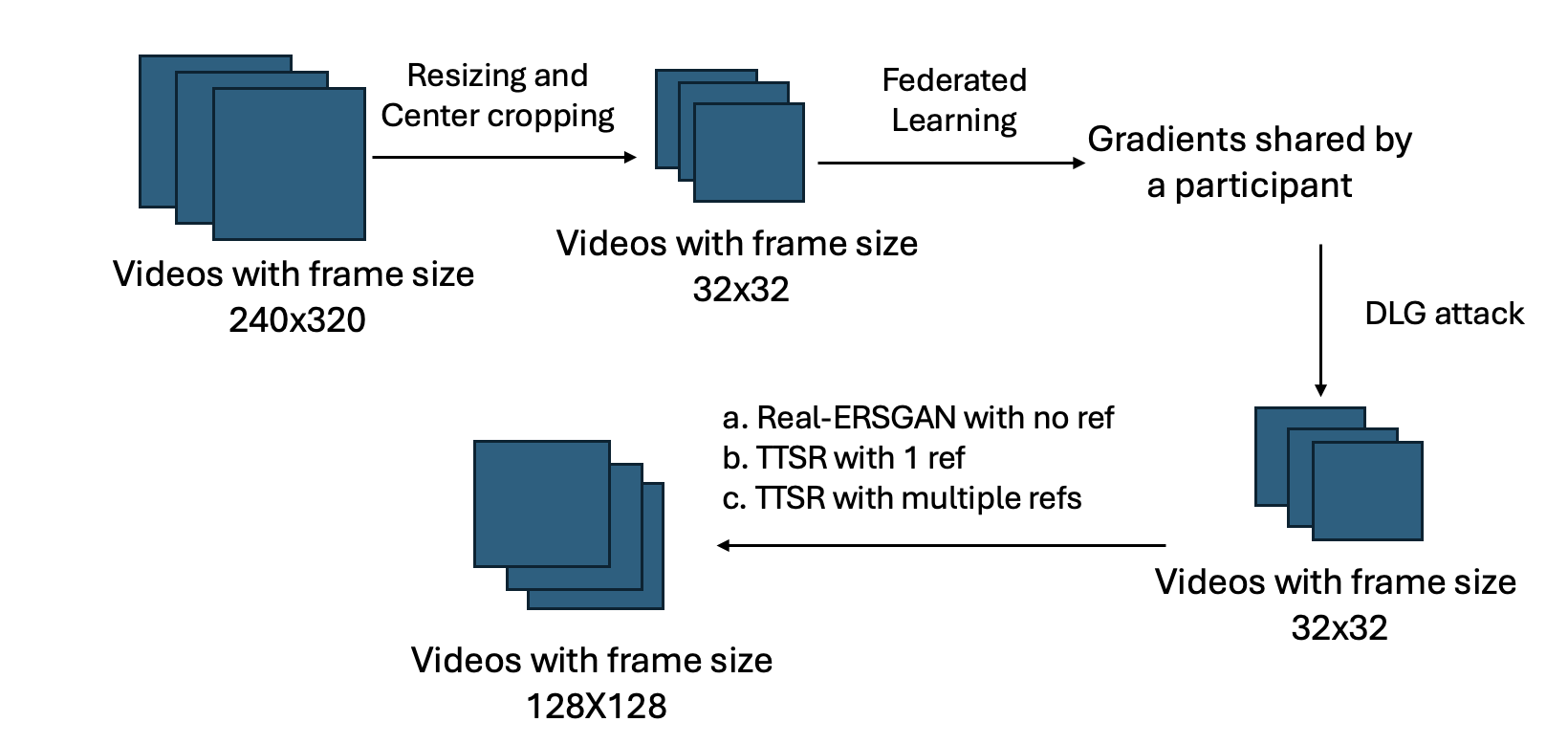}
    \captionsetup{justification=centering, width=\textwidth}
    \caption{Video Reconstruction Pipeline Without Neural Network Pre-processing.}
    \label{fig:without_preprocessing}
\end{figure*}

The initial step involves a transformation of the input video frames by resizing and center cropping. 
The frame size is reduced from $240$x$320$ to $32$x$32$ pixels with a batch size of 1. 
This downsampling step is crucial for applying the DLG attack, aligning with the requirements of this technique. 
The reduced-resolution video frames are then processed through a federated learning environment. 
In this context, gradients are extracted, which correspond to the video frames of batch size 1. 
This step simulates a federated training environment in which only gradient information is accessible. 
The video frames are reconstructed from the shared gradients using the DLG attack. 
The reconstructed video frames, now at a resolution of $32$x$32$, are subjected to super-resolution technique-based enhancement. 
To enhance the frame quality, we employed two distinct super-resolution techniques from \cite{wang2021realesrgan, yang2020learning}.  
For the scenario where the attacker does not have access to any video frame used to generate gradients, we incorporated Real-ESRGAN \cite{wang2021realesrgan}. 
For scenarios where the attacker has access to one or more frames, we utilized the \emph{Texture Transformer Network for Image Super-Resolution} (TTSR) \cite{yang2020learning}. 
Fig. \ref{fig:without_preprocessing} details our approach to video reconstruction in scenarios without feature extraction based on neural networks. 
\begin{itemize}
    \item \textbf{Real-ESRGAN with zero reference:} This method was used to enhance the frame resolution to 128x128 pixels without utilizing any external reference frames. 
    It improves blind super-resolution for real-world images through advanced degradation modeling and a U-Net discriminator, achieving superior restoration. 
    For our experiments, we used the open source implementation available on GitHub \href{https://github.com/xinntao/Real-ESRGAN-ncnn-vulkan}{https://github.com/xinntao/Real-ESRGAN-ncnn-vulkan}.

    \item \textbf{TTSR with one reference:} This approach utilizes a single reference frame to guide the super-resolution process, upscaling the frames to 128x128 pixels. 
    It leverages attention mechanisms for effective texture transfer from reference images. 
    We used the open-source TTSR implementation available on GitHub (\url{https://github.com/researchmm/TTSR}).
    
    \item \textbf{TTSR with multiple references:} This method extends the TTSR approach by leveraging multiple reference frames to refine the super-resolution reconstruction, also resulting in 128x128 pixel frames. 
\end{itemize}
\subsection{Gradient Inversion on Extracted Video Features}
\label{subsec:features}
We then explored the video reconstruction in the presence of neural network feature extraction. 
We used the research work on \emph{Collaborative Learning of Anomalies with Privacy (CLAP) } from \cite{al2024collaborative} as a use case.
This approach presents a privacy-preserving and collaborative learning paradigm specifically designed for unsupervised video anomaly detection. 
CLAP tackles the critical challenge of training models on sensitive and unlabeled surveillance video data, establishing a new baseline in this domain.
We utilized the open-source implementation available on GitHub (\url{https://github.com/AnasEmad11/CLAP}). 
We propose a systematic procedure that extends the concepts of DLG for the extraction of video frames in scenarios involving such pre-processing in Section \ref{sec:discussion}. 
Due to the restrictive effect of extensive video pre-processing on the efficacy of gradient inversion attacks, the intermediate feature matrices could not be retrieved, blocking the reconstruction of video frames. 
\subsection{Exploring the Effect of Feature Extractors on Gradient Inversion Attacks}
\label{subsec:effect_of_using_features}
To investigate the impact of feature extractors on gradient inversion attacks, we implemented three prominent attacks: \textbf{DLG, iDLG, and R-GAP}. 
The DLG and iDLG attacks are iteration-based recovery attacks, and R-GAP  \cite{zhang2022survey} is a recursion-based recovery attack. 

The \textbf{DLG} attack frames data reconstruction as an iterative optimization problem that begins with a random guess, $(\mathbf{x}', \mathbf{y}')$. 
The random data is then repeatedly updated using gradient descent to minimize the distance between its gradients and the victim's true gradients, eventually converging on the private data.
\begin{equation}
\label{eq1}
\begin{aligned}
\mathbf{x}'^*, \mathbf{y}'^* &= \arg\min_{\mathbf{x}', \mathbf{y}'} \left\| \nabla W' - \nabla W \right\|^2 \\
&= \arg\min_{\mathbf{x}', \mathbf{y}'} \left\| \frac{\partial\mathcal{L}(\mathcal{F}(\mathbf{x}', W), \mathbf{y}')}{\partial W} - \nabla W \right\|^2
\end{aligned}
\end{equation}

The core of the attack is formulated as the optimization problem shown in Eq. \ref{eq1}. 
The objective is to find the optimal reconstructed data $\mathbf{x}'^*$ and corresponding labels $\mathbf{y}'^*$ that best reproduce the leaked information. 
This is achieved by searching for the arguments $(\mathbf{x}', \mathbf{y}')$ that minimize the squared euclidean distance between two vectors: the ground-truth gradient leaked from the victim, $\nabla W$, and a synthetically generated gradient, $\nabla W'$.
The minimization is performed over the set of dummy inputs $\mathbf{x}'$ and $\mathbf{y}'$, which are initialized randomly and iteratively updated. 
The synthetic gradient $\nabla W'$ is calculated with respect to the model's weights $W$ by using the model's forward pass function, denoted $\mathcal{F}(\cdot)$, and its loss function, $\mathcal{L}(\cdot)$, evaluated on these dummy inputs.

A critical insight of the \textbf{iDLG} attack is that the ground-truth label, $c$ can be accurately recovered from the shared gradients by applying the heuristic shown in Eq. \ref{eq2}. 
This rule identifies the correct class index, $i$, by exploiting the property that its final-layer weight gradient, $\nabla \mathbf{W}_L^i$, is geometrically distinct, having a non-positive dot product with the gradients of all other incorrect classes 
\begin{equation}
\label{eq2}
    c = i, \quad \text{s.t.} \quad \nabla {\mathbf{W}_L^i}^T \cdot \nabla \mathbf{W}_L^j \le 0, \quad \forall j \ne i
\end{equation}
So, iDLG attack optimizes only the data and not the label, making the attack lot faster than DLG.

On the other hand, recursion-based attacks analytically recover data by recursively working backward through the network from the final layers to the input. \textbf{R-GAP} exploits the direct mathematical relationships between a layer's inputs, weights, and gradients; they solve for the input of each preceding layer until the original data is revealed \cite{zhu2020r}.  
This system directly relates a layer's input $x_i$ to its weights $W_i$ and feature map $Z_i$, allowing for its reconstruction by solving Eq.\ref{eq3}. 
The gradients of the loss with respect to the weights and the feature map are given by $\nabla W_i$ and $\nabla Z_i$, respectively.

\begin{equation}
\label{eq3}
\begin{aligned}
    W_i \cdot \mathbf{x}_i &= Z_i \\
    \nabla Z_i \cdot \mathbf{x}_i &= \nabla W_i
\end{aligned}
\end{equation}

An image dataset was selected for this analysis, as these attacks are demonstrably successful at reconstructing images from gradients. 
Our experiments proceeded in two stages. 
First, we applied the attacks to gradients generated directly from images using a simple and a moderately complex classifier, with the objective of reconstructing the original image. 
Second, we attacked gradients derived from feature vectors, which were generated by pre-trained extractors, to assess whether the original feature vectors could be successfully reconstructed. 
Fig. \ref{model_architectures} illustrates the architectures of the simple and moderately complex classifiers used in our experiments. 
The source code for our paper is available on GitHub\footnote{\url{https://github.com/fazal-029/VideoLeakage}}.

%% file: performance.tex
\section{Experiments}
\label{sec:performance}%
\begin{table*}[t!]
\centering
\captionsetup{justification=centering, width=\textwidth}
\caption{Quantitative Evaluation of Reconstructed Frame Fidelity for Different image super-Resolution Strategies.} 
\label{tab:performance} 

\resizebox{\textwidth}{!}{%
\begin{tabular}{|l|c|cc|cc|cc|}
\hline
& \multirow{3}{*}{\begin{tabular}[c]{@{}c@{}}Baseline (High vs. Low)\end{tabular}} & \multicolumn{2}{c|}{Video SR w/o Ref.} & \multicolumn{2}{c|}{Video SR w/ 1 Ref.} & \multicolumn{2}{c|}{Video SR w/ Mult. Refs.} \\
\cline{3-8}
& & Enhanced vs. High & Enhanced vs. Low & Enhanced vs. High & Enhanced vs. Low & Enhanced vs. High & Enhanced vs. Low \\
\hline
Video 1 & 28.40/0.2026 & \textbf{28.75}/0.2145 & 29.52/0.6103 & 28.68/0.2254 & 30.12/\textbf{0.6693} & 28.68/\textbf{0.2263} & \textbf{30.14}/0.6744 \\
\hline
Video 2 & 28.47/0.1968 & \textbf{28.86/0.2311} & 30.01/0.6842 & 28.76/0.2206 & \textbf{30.57}/\textbf{0.7006} & 28.76/0.2205 & \textbf{30.57}/0.6998 \\
\hline
Video 3 & 28.63/0.2881 & 28.74/0.2956 & 31.37/0.7978 & \textbf{28.95}/0.3177 & 32.76/0.8771 & \textbf{28.95/0.3178} & \textbf{32.77/0.8780} \\
\hline
\end{tabular}
}
\label{results}
\end{table*}
\textbf{Setup.}
\indent We utilized the UCF-Crime \cite{sultani2018real} and CIFAR-100 \cite{krizhevsky2009learning} datasets. 
The UCF-Crime dataset consists of 1,610 training videos and 290 testing videos, categorized into 13 distinct anomalous event classes. For our experiments, we used the I3D features extracted from the UCF-Crime dataset using an inflated 3D ResNet50 feature extractor consistent with the approach in the CLAP paper. 
The CIFAR-100 dataset contains 60,000 color images in 10 classes and 6000 images per class.
A ResNet20 model, pre-trained on 50\% of the CIFAR-100 dataset, was employed as a feature extractor with its final layer removed.
The experiments were conducted within the PyTorch framework \cite{paszke2019pytorch}, configured to run on either a CPU or an available NVIDIA CUDA GPU. 
The optimization was performed using the L-BFGS optimizer.
To ensure the reproducibility of our results across all experimental runs, a fixed manual seed was set for the PyTorch random number generator. 
\subsection{Results on Raw Video Frames with Simple Transformations}
\label{subsec:results_on_raw_frames}
\begin{figure*}[htbp]
    \centering
    \includegraphics[width = 0.65\textwidth]{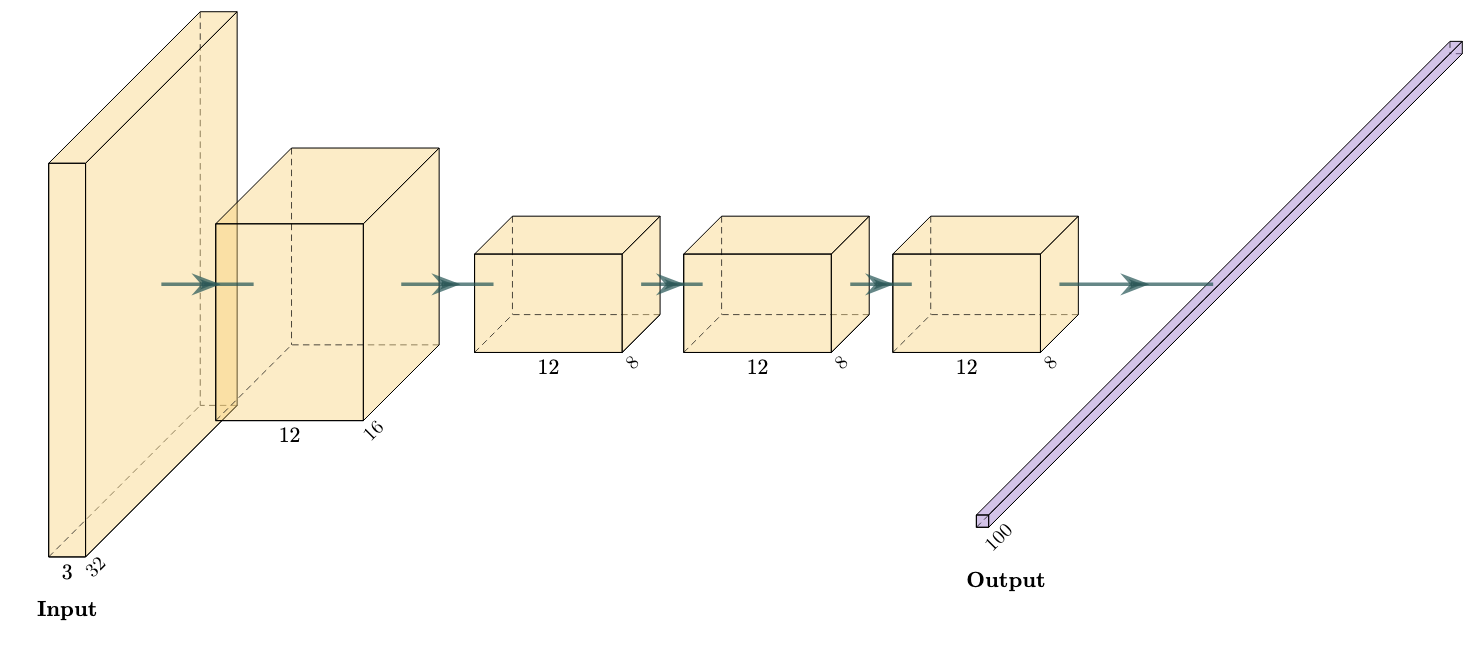}
    \captionsetup{justification=centering, width=\textwidth}
    \caption{LeNet used in the DLG attack.}
    \label{fig:lenet}
\end{figure*}

Our main goal was to investigate the extensibility of gradient inversion attacks on video data. 
We initially approached videos as a sequence of individual frames. 
Three videos were selected from the UCF crime dataset, each with distinct levels of camera movement: static (Video 2), minimal (Video 1), and frequent (Video 3). 
The original 240x320 resolution of the video frames was downsampled to 32x32. 
A batch size of 1 was used for processing. We simulated a federated learning environment to model the sharing of gradients for training a global model. 
For the experiments, we employed a LeNet classifier \cite{lenet}, mirroring the methodology of the DLG attack \cite{zhu2019deep}. 
The specific architecture, as detailed in the original DLG study, is illustrated in Fig.\ref{fig:lenet}. 
The block diagrams were prepared using the PlotNeuralNet software package \cite{iqbal2018plotneuralnet}.   
By applying the DLG attack to these shared gradients, we successfully reconstructed the original training frames. 

\begin{figure*}[h]
    \centering
    \begin{subfigure}[t]{\textwidth}
        \centering
        \begin{subfigure}[b]{0.35\textwidth}
            \includegraphics[width=\linewidth, height=5cm]{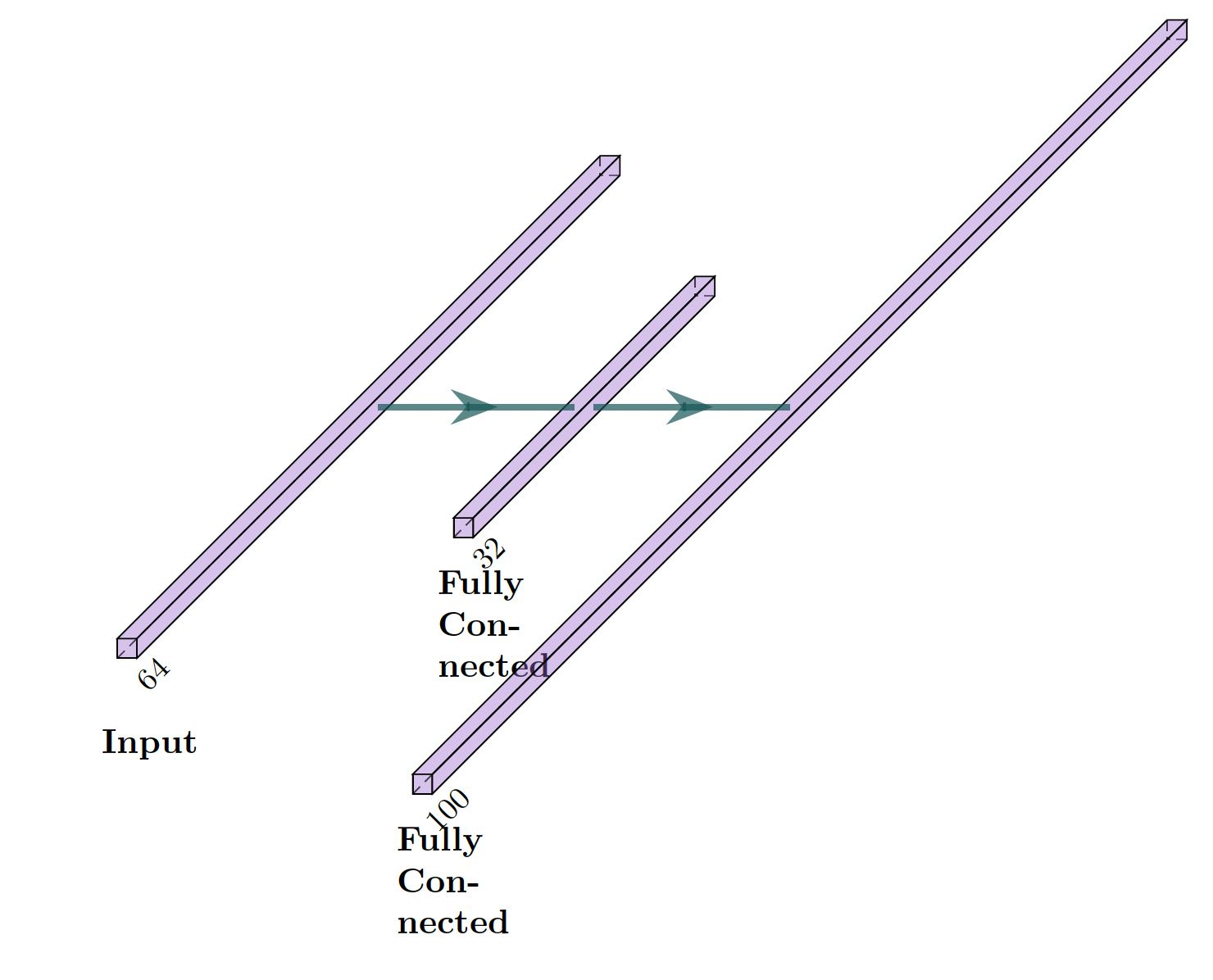}
        \end{subfigure}
        \hfill
        \begin{subfigure}[b]{0.55\textwidth}
            \includegraphics[width=\linewidth, height=5cm]{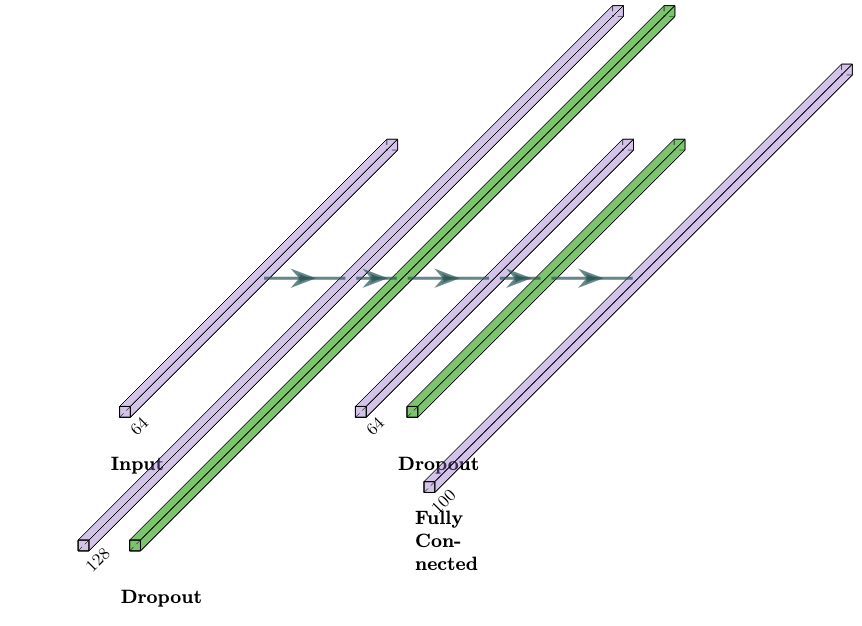}
        \end{subfigure}
    \end{subfigure}
    
    \caption{Architectures of the simple classifier (left) and the moderately complex classifier (right) used in our experiments.} 

    \label{model_architectures}
    
\end{figure*}

\begin{figure*}[htbp]
    \centering
    
    \begin{subfigure}[t]{\textwidth}
        \centering
        \begin{subfigure}[b]{0.32\textwidth}
            \includegraphics[width=\linewidth, height=3cm]{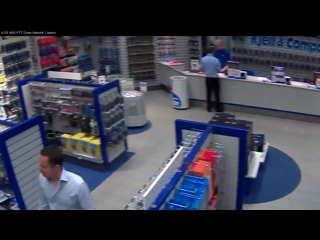}
        \end{subfigure}
        \hfill
        \begin{subfigure}[b]{0.32\textwidth}
            \includegraphics[width=\linewidth, height=3cm]{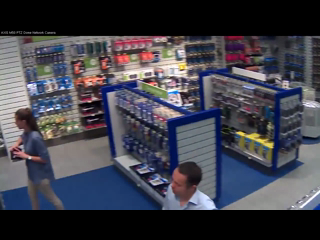}
        \end{subfigure}
        \hfill
        \begin{subfigure}[b]{0.32\textwidth}
            \includegraphics[width=\linewidth, height=3cm]{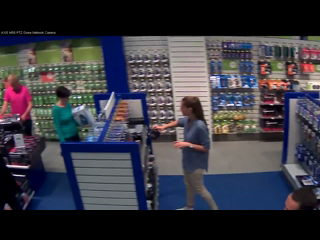}
        \end{subfigure}
        \caption{Representative video frames illustrating scenes from the UCF-crime dataset.} 
        \label{fig:row1}
    \end{subfigure}
    
    \vspace{0.25cm} 
    
    \begin{subfigure}[t]{\textwidth}
        \centering
        \begin{subfigure}[b]{0.32\textwidth}
            \includegraphics[width=\linewidth, height=3cm]{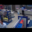}
        \end{subfigure}
        \hfill
        \begin{subfigure}[b]{0.32\textwidth}
            \includegraphics[width=\linewidth, height=3cm]{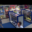}
        \end{subfigure}
        \hfill
        \begin{subfigure}[b]{0.32\textwidth}
            \includegraphics[width=\linewidth, height=3cm]{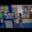}
        \end{subfigure}
        \caption{Representative video frames used as input for training the model.} 
        \label{fig:row2}
    \end{subfigure}
    
    \vspace{0.25cm} 
    
    \begin{subfigure}[t]{\textwidth}
        \centering
        \begin{subfigure}[b]{0.32\textwidth}
            \includegraphics[width=\linewidth, height=3cm]{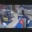}
        \end{subfigure}
        \hfill
        \begin{subfigure}[b]{0.32\textwidth}
            \includegraphics[width=\linewidth, height=3cm]{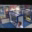}
        \end{subfigure}
        \hfill
        \begin{subfigure}[b]{0.32\textwidth}
            \includegraphics[width=\linewidth, height=3cm]{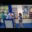}
        \end{subfigure}
        \caption{Sample video frames obtained via the DLG attack.} 
        \label{fig:row3}
    \end{subfigure}
    
    \vspace{0.25cm} 
    
    \begin{subfigure}[t]{\textwidth}
        \centering
        \begin{subfigure}[b]{0.32\textwidth}
            \includegraphics[width=\linewidth, height=3cm]{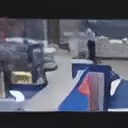}
        \end{subfigure}
        \hfill
        \begin{subfigure}[b]{0.32\textwidth}
            \includegraphics[width=\linewidth, height=3cm]{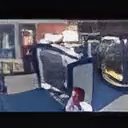}
        \end{subfigure}
        \hfill
        \begin{subfigure}[b]{0.32\textwidth}
            \includegraphics[width=\linewidth, height=3cm]{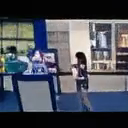}
        \end{subfigure}
        \caption{Sample enhanced video frames produced by the no-reference super-resolution method.} 
        \label{fig:row4}
    \end{subfigure}

    \begin{subfigure}[t]{\textwidth}
        \centering
        \begin{subfigure}[b]{0.32\textwidth}
            \includegraphics[width=\linewidth, height=3cm]{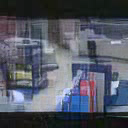}
        \end{subfigure}
        \hfill
        \begin{subfigure}[b]{0.32\textwidth}
            \includegraphics[width=\linewidth, height=3cm]{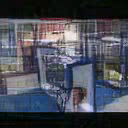}
        \end{subfigure}
        \hfill
        \begin{subfigure}[b]{0.32\textwidth}
            \includegraphics[width=\linewidth, height=3cm]{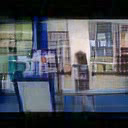}
        \end{subfigure}
        \caption{Sample enhanced video frames produced by super-resolution with one reference.} 
        \label{fig:row5}
    \end{subfigure}

    \begin{subfigure}[t]{\textwidth}
        \centering
        \begin{subfigure}[b]{0.32\textwidth}
            \includegraphics[width=\linewidth, height=3cm]{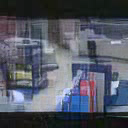}
        \end{subfigure}
        \hfill
        \begin{subfigure}[b]{0.32\textwidth}
            \includegraphics[width=\linewidth, height=3cm]{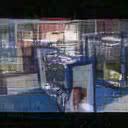}
        \end{subfigure}
        \hfill
        \begin{subfigure}[b]{0.32\textwidth}
            \includegraphics[width=\linewidth, height=3cm]{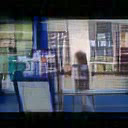}
        \end{subfigure}
        \caption{Sample enhanced video frames produced by super-resolution with multiple references.} 
        \label{fig:row6}
    \end{subfigure}
    
    \caption{Frames obtained through DLG and enhanced using super-resolution.} 
    \label{fig:all_rows}
\end{figure*}

The optimization for each frame was run for 300 iterations.
A frame's reconstruction was considered successful if the loss, the sum of the squared difference between original gradients and calculated gradients, fell below a threshold of 0.00001. 
If this threshold was not met, the process was re-initialized and attempted up to a maximum of ten times.
We then applied the image super-resolution techniques described in Section \ref{subsec:raw_video_frames}. 
We used Real-ESRGAN to simulate the scenario where the attacker does not have access to any frames used in training. 
We used TTSR with one reference frame from the video to account for the scenario where the attacker has only one frame from the video, and finally, we used TTSR with multiple reference frames to simulate the scenario where the attacker has access to more than one video frame. 
All three image super-resolution approaches upscale the frame resolution from 32x32 to 128x128. 
Output videos were generated and saved with a frame rate of 30.
Fig. \ref{fig:without_preprocessing} presents a visualization of the steps followed here.

We evaluated video quality enhancement using Peak Signal-to-Noise Ratio (PSNR) and Structural Similarity Index (SSIM) as primary metrics, computed frame-wise across videos resized to 128×128 pixels for standardized comparison.
PSNR measures the ratio between the maximum possible power of a signal and the power of corrupting noise, commonly serving as an objective indicator of reconstruction quality in lossy image and video compression. 
For images with a bit depth of 8, typical PSNR values range from 30 to 50 dB \cite{fowler2025} with higher values generally correlating with reduced distortion. 
But the PSNR correlation with perceived visual quality is limited as noted in \cite{wang2004image}. 
SSIM, on the other hand, measures image reconstruction quality by evaluating changes in structural information, luminance, and contrast between the original and reconstructed images. 
SSIM values closer to 1 indicate higher similarity and better visual quality as noted in \cite{wang2004image}.

Table \ref{results} presents a quantitative analysis of the reconstructed frame quality achieved by different approaches. 
The terms used in Table \ref{results} are defined as follows:
\begin{itemize}
    \item Enhanced: Video frames resulting from the application of image super-resolution to the frames reconstructed by deep leakage attack (image super-resolution without reference, with one reference, and with multiple references).
    \item High: The original UCF Crime Videos.
    \item Low: Video frames reconstructed by deep leakage attack before the application of image super-resolution.
\end{itemize}

To ensure compatibility with PSNR and SSIM metrics, which require frames of identical resolution, all frames under examination were converted to a uniform resolution of 128x128 pixels. 
The baseline results (High vs. Low quality) indicate that DLG attack can effectively reconstruct video frames from gradients only. image super-resolution (SR) techniques further refine these reconstructed frames, increasing their utility for an attacker.
Fig. \ref{fig:all_rows} shows a few sample frames obtained through DLG and enhanced using super-resolution techniques. 
For image super-resolution without a reference frame, maximum quality enhancement was observed primarily in scenarios with minimal or no camera movement. 
In contrast, when dealing with frequent camera movement, employing multiple reference frames yielded the most substantial performance improvements. 
The experimental results demonstrate that the application of image super-resolution techniques leads to a measurable improvement in the structural similarity of the reconstructed videos, as quantified by SSIM.
\begin{figure*}[htbp]
    \centering
    \includegraphics[width =0.75\textwidth, keepaspectratio]{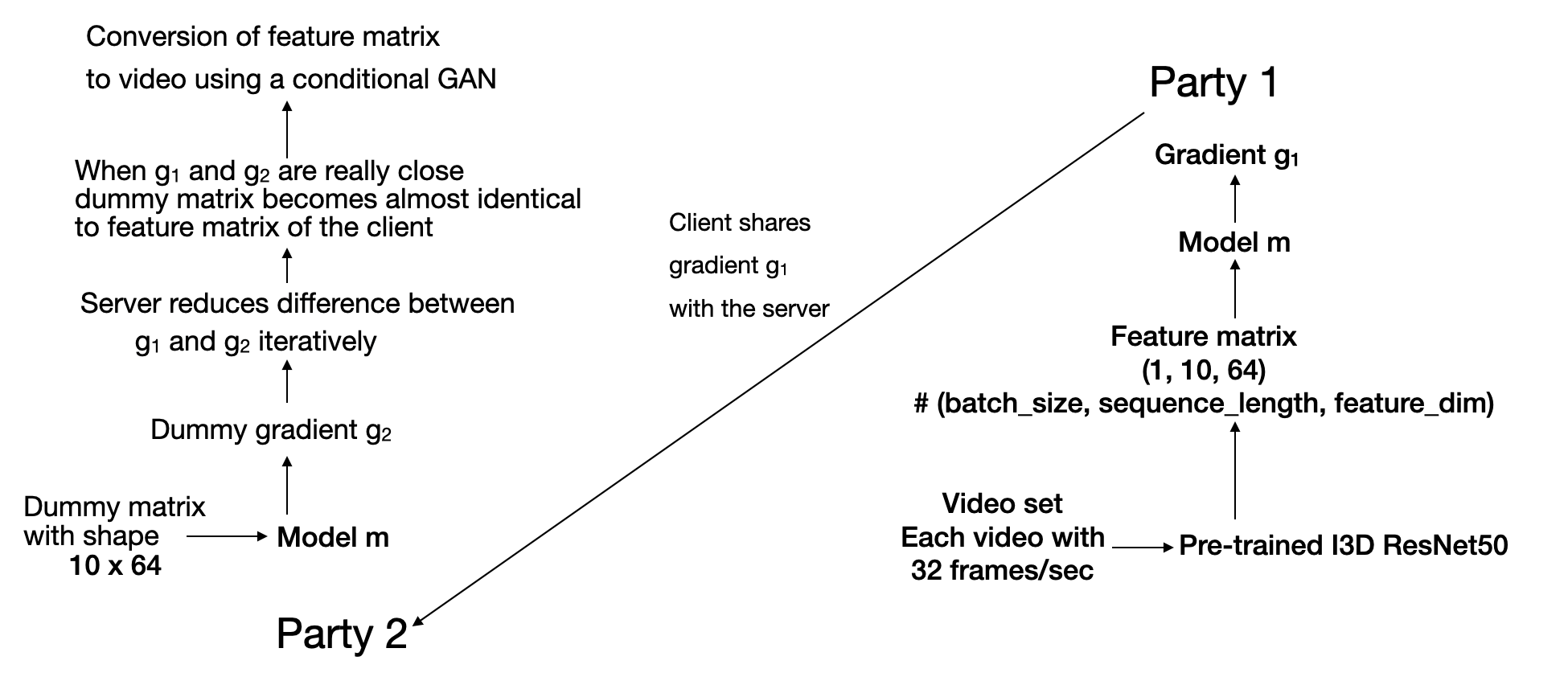}
    \captionsetup{justification=centering, width=\textwidth}
    \caption{Gradient Leakage Attack: Proposed Video Reconstruction Steps via Conditional GAN.}
    \label{fig:with_preprocessing}
\end{figure*}
\subsection{Results of Gradient Inversion Attacks on Extracted Video Features}
\label{results_on_features}
\begin{figure*}[htbp]
    \centering
    \includegraphics[width = 0.75\textwidth]{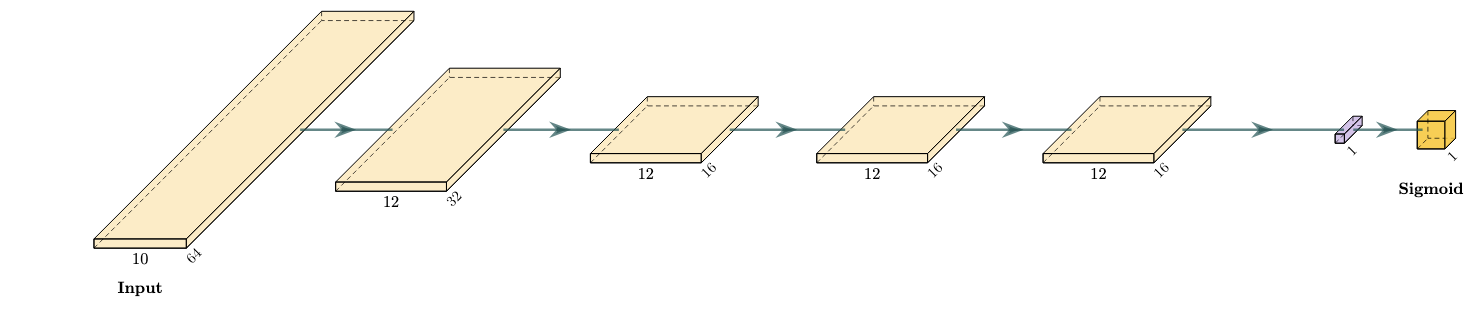}
    \captionsetup{justification=centering, width=\textwidth}
    \caption{Modified LeNet-style classifier.}
    \label{fig:lenet_modified}
\end{figure*}

For the second video processing approach, where videos are first fed to a pre-trained feature extractor, we used the readily extracted I3D features from the GitHub repository \href{https://github.com/tianyu0207/RTFM}{https://github.com/tianyu0207/RTFM}.
It provides feature matrices of shape (1, 10, 2048) for each clip.
To meet the operational requirements of the DLG, we applied max pooling to downsample these matrices to a shape of (1, 10, 64). 
These downsampled feature matrices were then utilized in a federated learning framework. 
The DLG algorithm was subsequently applied to the gradients associated with these matrices to determine if the original feature matrices could be successfully reconstructed. 
Fig. \ref{fig:with_preprocessing} presents our proposed steps for reconstructing videos for the second video handling approach.
We used a modified LeNet-style classifier depicted in Fig.  \ref{fig:lenet_modified} that takes these I3D features as input. It is similar to the LeNet architecture, known for its vulnerability to the DLG attack. 
This time, the DLG attack was executed for 20,000 iterations. 
To mitigate convergence to local minima, the loss was monitored every 1,000 iterations. 
If stagnation was detected, noise sampled from a normal distribution ($\mu=0$, $\sigma=0.001$) was introduced to the dummy data, and the optimizer's learning rate was halved. 
We tried with both Adam and L-BFGS optimizers.
But the DLG attack failed to reconstruct the feature matrices accurately this time. 
\subsection{Influence of Feature Extractors}
Finally, to evaluate the effect of the feature extractor on gradient inversion attacks, we selected three established algorithms: DLG, iDLG, and R-GAP. 
We chose the CIFAR-100 dataset \cite{krizhevsky2009learning} for this experiment, given its documented susceptibility to image reconstruction by these attacks.
A ResNet20 model, trained on 50\% of the CIFAR-100 dataset, was used as the feature extractor after removing its final layer. 
We then performed two distinct experiments within a federated learning framework. 
In these experiments, the feature extractor was paired with either a simple or a moderately complex classifier (Fig. \ref{model_architectures}) to assess whether gradient inversion attacks could successfully reconstruct the intermediate feature matrices from the gradients. 
We recorded our findings in Table \ref{results_feature_extractor}
\begin{table*}[t!]
\centering
\captionsetup{justification=centering, width=\textwidth}
\caption{Reconstruction Performance of DLG, iDLG, and R-GAP on Feature Matrices with Varying Classifier Complexity (N - feature matrices could not be retrieved from gradients, Y - feature matrices were retrieved from gradients.)} 
\label{tab:dlg_idlg_r-gap} 
\resizebox{\textwidth}{!}{%
\begin{tabular}{|c|c|c|c|c|}
\hline
 & \begin{tabular}[c]{@{}c@{}}Moderately complex \\      classifier with feature extractor\end{tabular} & \begin{tabular}[c]{@{}c@{}}Moderately complex \\      classifier with raw images\end{tabular} & Simple classifier with feature   extractor & Simple classifier with raw images \\ \hline
DLG & N & Y & Y & Y \\ \hline
iDLG & N & Y & Y & Y \\ \hline
R-GAP & N & N & N & N \\ \hline
\end{tabular}
}
\label{results_feature_extractor}
\end{table*}
The results in Table \ref{results_feature_extractor} demonstrate that utilizing feature matrices does not, by itself, prevent gradient inversion attacks, as these attacks were successful when a simple classifier was employed.
Notably, while moderately complex classifiers are vulnerable to data leakage from gradients when trained on raw images, classifiers of the same complexity prevent such leakage when a feature extractor is used. 
This was demonstrated in our experiment, where a standard LeNet model was susceptible to attack, yet a modified LeNet of comparable complexity became robust when operating on extracted features.
Therefore, we conclude that the inclusion of a feature extractor significantly increases the difficulty of executing gradient inversion attacks.

%% file: discussion.tex
\section{Discussion}
\label{sec:discussion}
Our investigation into gradient inversion attacks across two video processing scenarios revealed that DLG is ineffective at reconstructing pre-processed feature matrices from their corresponding gradients. 
A sharp contrast was observed between the two video processing approaches.
In our frame-wise approach, the attack perfectly recovered the original image sequence, resulting in a total leakage of temporal information. 
Conversely, for our feature matrix approach, the attack failed entirely.
However, subsequent experiments with various attack models on the CIFAR-100 dataset demonstrated that while employing a feature extractor does not completely prevent gradient inversion, it does introduce a significant layer of complexity to the attack process.
Pre-processing with a feature extractor and conversion to matrices pose some unique challenges for video data. 
Evaluating the quality of the reconstructed feature matrices cannot be achieved through direct visual inspection or linguistic evaluation, as it is possible with images or texts. 
Also, unlike images, which exhibit strong spatial correlations captured in gradients due to the similarity of neighboring pixels, and text, where sequential dependencies between words are often encoded in gradients, feature matrices lack inherent spatial or sequential structure. 
This absence of discernible patterns in feature matrices significantly complicates gradient inversion. 
Although the existing literature demonstrates tremendous success for image and text, gradient inversion attacks on numerical feature matrices are still an active area of research. 
The most pertinent work is TabLeak  \cite{vero2022tableak}, which achieves state-of-the-art performance in tabular data containing numerical and categorical features. 
Notably, TabLeak's success with tabular data relies heavily on the presence of categorical columns in the data vector. 
Their results indicate a significant performance gap, with continuous features exhibiting up to 30\% lower reconstruction accuracy compared to categorical features for the same batch size. 
However, our scenario involves feature matrices consisting exclusively of continuous numerical features, a specific case not addressed in the existing literature, including TabLeak.

%% file: conclusion.tex
\section{Conclusion}
\label{sec:conclusion}
\balance
This paper examines the vulnerability of video data to deep leakage attacks based on shared gradients in different pre-processing scenarios. 
We investigate the retrieval and utility of the training video frames obtained through such attacks and explore the potential of image super-resolution to improve the quality of reconstructed frames.
Our experiments show that image super-resolution techniques improve the frame quality.
Recent advancements in gradient inversion attacks, as reported in the literature, reveal their growing effectiveness even with high-resolution imagery and larger batch sizes. 
This reduces the safety of collaborative systems, previously deemed safe, where the safety depends on either high-resolution images or increased batch size.
Notably, our findings indicate that employing a pre-trained feature extractor offers a promising strategy for mitigating gradient inversion attacks like DLG in federated learning settings. 
However, even in such systems, the configuration is of importance, and as such, simple configurations can leak data.
Our future work will, therefore, concentrate on a comprehensive investigation into this issue.
%
And while this work demonstrates that feature extractors provide resilience, our crucial next step is to investigate the causes. 
Future work will focus on isolating the specific factors contributing to the attack's failure. 
Finally, we plan to extend our analysis to more contemporary attack strategies, such as those leveraging generative models or attacks on compressed gradients. 
%